\documentclass[twoside]{article}

\usepackage{aistats2024}
\RequirePackage{algorithm}
\RequirePackage{algorithmic}
\usepackage{graphicx}
\usepackage{booktabs} 
\usepackage{amsmath}
\usepackage{amssymb}

\RequirePackage{natbib}
\RequirePackage{eso-pic} 
\RequirePackage{forloop}
\RequirePackage{url}

\begin{document}

%

%

\twocolumn[

\aistatstitle{Shapley Marginal Surplus for Strong Models}

\aistatsauthor{ Daniel de Marchi \And Michael Kosorok \And  Scott de Marchi }

\aistatsaddress{ University of North Carolina \And  University of North Carolina \And Duke University } ]

\begin{abstract}

Shapley values have seen widespread use in machine learning as a way to explain model predictions and estimate the importance of covariates. Accurately explaining models is critical in real-world models to both aid in decision making and to infer the properties of the true data-generating process (DGP). In this paper, we demonstrate that while model-based Shapley values might be accurate explainers of model predictions, machine learning models themselves are often poor explainers of the DGP even if the model is highly accurate. Particularly in the presence of interrelated or noisy variables, the output of a highly predictive model may fail to account for these relationships. This implies explanations of a trained model's behavior may fail to provide meaningful insight into the DGP. In this paper we introduce a novel variable importance algorithm, Shapley Marginal Surplus for Strong Models, that samples the space of possible models to come up with an inferential measure of feature importance. We compare this method to other popular feature importance methods, both Shapley-based and non-Shapley based, and demonstrate significant outperformance in inferential capabilities relative to other methods.

\end{abstract}

\section{Problem Introduction}

With the rise in popularity of difficult-to-interpret "black box" models, Shapley values have become a common tool in the growing field of explainable AI \citep{burkart2021survey}. Originally developed for trees, and later generalized to any model, these algorithms provide explanations for the outputs of a trained model using the Shapley values of the covariates with respect to each individual prediction \citep{lundberg2020local, lundberg2017unified}. Researchers can gain a precise understanding of how each variable influences the predictions of their model. This is especially critical for deployed systems that will interact with sensitive data, or cases where models are not the primary decision makers and their outputs need to be explained to the primary decision maker. For instance, if the model is predicting the success or failure of a drug treatment for a specific patient, a clinician might want to review the model's decision and make sure that the factors leading to the model's output aligns with their medical knowledge. However, there are significant limitations to variable importance algorithms based on a single supervised learning model. These algorithms assume that either this model is a reasonable characterization of the DGP, or that the DGP is not important in the current application \citep{kumar2020problems}. 

The example frequently used in the literature to justify focusing on model-based explanations is a bank with a highly accurate loan granting model. The bank may want explanations of this model so that they can recommend strategies to their customers to be approved for a loan. However, the actual validity of the model isn't relevant to the problem. If the true variable responsible for a customer's untrustworthiness for a loan is their income, but the bank's model denied them based on their credit score, this is not an issue to the bank since their primary goal is model accuracy, not inference. The two variables are likely tightly correlated and the model is highly accurate, so whether or not the credit score is actually part of the DGP isn't an issue to the bank \citep{chen2020true}.

In many cases however, researchers need models that are accurate with respect to the true DGP. If a researcher is testing a model for racial bias, it is critical that the model is an accurate representation. If other variables (such as where a person lives, their medical history, etc.) are associated with an individual's race, the model may not use race as a feature even though if it is part of the true DGP, and racial bias exists in the data. A feature importance tool that focuses on a single model may give the illusion of no racial bias even though the underlying data is actually highly racially biased \citep{chen2020true, kumar2020problems}. Some feature importance algorithms attempt to solve this problem by fitting multiple models to consider a variety of different possible explanations, depending on the class of models chosen.

The Model Class Reliance (MCR) algorithm approaches this by conducting an analysis on the "Rashomon set" of relatively strong models. It fits a large number of these high-quality models, then averages the permutation importance of each feature to the Rashomon set of strong models. While this method is not directly Shapley based, it works off similar principles, but has additional useful properties due to assessing feature importance across a broad class of possible models \citep{fisher2019all}. 

Leave-One-Covariate-Out (LOCO) uses the loss of a full model relative to submodels that have been refit without each covariate to evaluate the marginal importance of each covariate \citep{lei2018distribution}. Typically, this requires generating a number of train/test sets from an original dataset, and assessing the LOCO importance across these different replicates.

While these approaches provide better explanations of the true DGP than single-model approaches, significant weaknesses exist. These include failing to consider the full class of models, failing to correctly analyze the full class of models, and failing to properly make use of Shapley values. To this end, we introduce Shapley Marginal Surplus for Strong Models (SMSSM) to generate explanations that respect the true underlying DGP.

\section{Theoretical Preliminaries}

Consider a supervised model $f \in \mathbb{F}$ for some class of possible models $\mathbb{F}$, where we are training on data $X$ and target variable $y$, such that our predicted $\hat{y} = f(X), X = [X_1...X_p]$. Denote any individual observation of $X$ to be $x \in X$. 

\textbf{Definition} Data-Generating Process (DGP): The data-generating process of a dataset is the true functional relationship between $y$ and $X$, and can be expressed as $y \sim F(X) + \epsilon(X)$, where $\epsilon(X)$ is some irreducible error distribution that may or may not depend on the input data. In general, outside of simulated data this true relationship is unknown.

Classical statistical inference is concerned with identifying this true relationship: a common approach is to assume that $F$ is a linear function of the features, $\epsilon$ does not depend on $X$, and to then conduct hypothesis tests about the parameters of the realized model $f$. Coefficients that are not statistically significantly different from zero are assumed to be not present in the true functional form $F$. 

These hypothesis tests are conducted under the assumption that the true function $F(X)$ may only use a subset of the features. Out of our total subset of features $X = [X_1...X_p]$, statistical hypothesis tests attempt to identify some subset $T = [t_1...t_k] \subseteq [1, 2, ... p], X_T = [X_{t_1}...X_{t_k}] \subseteq [X_1...X_p]$ such that the elements in $X_T$ are the elements in the true functional form of $F(X)$. This gives us the following property: 

\[F(X_T) = F(X) \ \forall \ x \ \in \ X\]

Note that $T$ is the smallest such set that the above holds. In fact, for any set of covariates $W$ such that $F(X_W) = F(X) \forall x \in X$, we know that $T \subseteq W$.

\textbf{Definition} Feature Importance: A feature importance is an allocation of weights $w = [w_1...w_p]$, where the weight indicates the change in either a quality metric or the model's predictions when the corresponding feature $X_i$ is removed relative to the original predictions. The sum is sometimes normalized such that $\sum_{i=1}^p w_i = 1$. Some possible examples of quality functions are mean squared error (MSE) for regression or the area under the receiver operating characteristic (ROC) curve for a binary classification problem \citep{covert2021explaining}.

Feature importance can be analyzed locally on a per-prediction level for each $x \in X$ or globally across the entire dataset $X$ \citep{lundberg2017unified}. Additionally, we can distinguish between selective feature importance and attributive feature importance. 

\textbf{Definition} Attributive and Selective Importance: we define a feature importance function to be $selective$ if the weight vector $w$ can be used to identify the subset of true covariates $X_T$ by analyzing the weights where $w_i > 0$, or by conducting hypothesis tests on the realized distribution of the components of sampled weight vectors $w^1...w^k$ for some sampling method. This makes a selective feature importance function more similar to a statistical hypothesis test.

The alternative is attributive feature importance. Attributive feature importance is concerned with explaining the properties of the model $f$, and not the true data-generating process $F(X)$, so a positive weight only implies that the feature is useful to that particular model. The assumption for an attributive variable importance analysis is that either $f(X) \sim F(X)$ to a sufficient degree or that we are not concerned with the true relationship and are only concerned with the trained model.

We are making this distinction so we can later discuss whether a method is specific to a particular model or if it is an inferential technique about the underlying mechanism. We also note that selective explanation methods can still vary in the explanation returned and may depend on the functional class chosen: for instance, the explanation of a random forest and a gradient boosted ensemble might be dramatically different. However, both explanations will still only be based on covariates in the true-data generating process.

\section{Subset Functions}

To analyze how a model performs when a particular feature is removed, we need to create a way to generate functions that will behave similarly to the trained model $f$, but only operate on an arbitrary subset of the  features used by $f$. This is known as a subset function \citep{covert2021explaining}.

Consider a set $\mathbb{P}(p)$ of every possible binary vector of length $p$. Because $X$ is $p$ dimensional, any subset of the covariates in $X$ can be expressed as $X \times s$ for some $s \in \mathbb{P}(p)$, where $X = [X_1...X_p]$ and $s$ is a vector of zeros and ones.

\textbf{Definition} Subset Function: A subset function is a function that models $g(X, s) \sim y$, such that covariates excluded by the indicator vector $s$ are not present in the functional form of $g$. That is, we have $g(x, s) = g(x', s)$ for all triplets $(x, x', s)$ such that $x \times s = x' \times s$. 

\textbf{Definition} Subset Extension: a subset extension of $f$ is a subset function $f_E$ such that $f(x) = f_E(x, s)$ whenever $s$ is a vector of only ones, i.e. in the presence of all features. Therefore the model $f$ and $f_E$ satisfy:

\[f(x) = f_E(x, [1, 1, .... 1]) \ \forall \ x \ \in \ X\]

There are many approaches to creating subset functions, which we will not cover exhaustively. The four broad groups of methods are as follows:

\begin{itemize}
    \item Constant value replacement: for any $s_i$ equal to 0, all values in the corresponding column $X_i$ are replaced with a constant value $c$, typically either the expected value or zero.
    \item Distribution-based replacement: for any $s_i$ equal to 0, all values in the corresponding column $X_i$ are replaced using a probability distribution $p_i$, such that $X_i \sim p_i(X \times s)$. Typically, this is the conditional distribution of $X_i$ on the non-removed values of $X$.
    \item Model-based replacement: for any $s_i$ equal to 0, all values in the corresponding column $X_i$ are replaced using an imputation model such that $X_i \sim g_i(X \times s)$ for some arbitrary imputation model $g_i$.
    \item Refit-based replacement: this refits the model entirely to not use any columns $X_i$ such that $s_i = 0$, and analyzes performance relative to the original.
\end{itemize}

The first three methods can be understood as function-based replacements, where each value in the column $X_i$ is replaced by a value $h_i(x, s)$ for some function of the non-removed covariates. These methods constrain that $\exists X_i \ s.t. \ h_i(X, s) \neq X_i$, because otherwise no values have been "removed" in any meaningful sense.

All four of these approaches can satisfy the definitions of a subset extension, and have varying benefits, drawbacks, and computational requirements. Now that we have defined some of the terminology for how to simulate the removal of features from a model, we turn to a brief discussion of game theory, and how game-theoretic concepts are used to calculate feature importance weights from the changes in model performance when features are removed using a subset extension method \citep{covert2021explaining}.

\section{Game Theory and Shapley Values}

\textbf{Definition} Cooperative Game: a cooperative game is mapping between a set of actors $a \in A$, where $A = [1...n]$, and a set function $\nu$, where $\nu(a)$ is the reward an individual set of actors would receive from playing the game in question. Often, the reward is a step function, where a certain quota needs to be met for a group of actors to receive a positive reward. For this section, assume that any natural integer $1 \leq i \leq n$ is an actor $i \in A$, and an arbitrary group of actors $[a_1, a_2, ... a_j]$ are the members of our sets $a \in A$ \citep{branzei2008models}. 

$\nu(\emptyset)$ is typically $0$ for any cooperative game, and the nonzero winnings are typically normalized such that the maximum equals 1. If we wanted to divide these winnings between players based on their pivotality to receiving rewards, we would turn to the Shapley value. The Shapley value is a way to allocate the surplus (reward) between the team such that each actor is compensated for their pivotality in the game. It is calculated from the set of possible marginal utilities.

\textbf{Definition} Marginal Utility: in a cooperative game, the marginal utility of player $i$ to a set $a \in A$, where $i \in a$ is $\nu(a) - \nu(a / \{i\})$, i.e. the reward of the set $a$ with player $i$ minus the reward without player $i$.

\textbf{Definition} Shapley Value: Label the payouts to each player for this coalition $\phi_1...\phi_j$, where each value is calculated as follows:

\[\phi_i = \sum_{a \subseteq A/\{i\}]} \frac{|a|! (n-|a| - 1)!}{n!}(\nu(a \cup \{i\}) - \nu(a))\] 

This can be interpreted as the marginal contribution of player $i$ to all possible subsets of $A$ that do not already include player $i$, normalized by the player count. These $\phi_j$ values are the Shapley values for each player. Shapley values have several useful properties that are applicable to variable importance \citep{shapley1953value, lundberg2020local}. 

\begin{itemize}
    \item Efficiency: the sum of Shapley values of individual players add up to the reward of the grand coalition minus the reward of the empty coalition.

    \[\sum_{i=1}^n \phi_i = \nu([1...n]) - \nu(\emptyset)\]
    
    \item Symmetry: two players with identical marginal contributions to all coalitions that do not include either of them have the same Shapley value.

    \[\text{If }\forall a: i, j \notin a, \ \nu(a \cup \{i\}) = \nu(a \cup \{j\}) \ \to \phi_i = \phi_j\]
    
    \item Monotonicity: if one player has greater contribution than another player to all coalitions that contain neither of them, then that player has the greater Shapley value.

    \[\nu(a \cup \{i\}) \geq \nu(a \cup \{j\}) \ \forall \ a \to \phi_i \geq \phi_j\]
    
    \item Null Player: a player with zero marginal contribution for all coalitions has a Shapley value of zero.

    \[\nu(a \cup \{i\}) = \nu(a) \ \forall \ a \to \phi_i = 0\]
    
    \item Additivity: a player's Shapley value for two independent games $G_1, G_2$ is the sum of their Shapley value for each individual game.

    \[\phi_{i \in G_1, G_2} = \phi_{i \in G_1}+ \phi_{i \in G_2}\]
\end{itemize}

The Shapley value is the only formulation that satisfies all these properties \citep{shapley1953value}. Shapley values can account for a feature's contribution even in the face of complex interrelationships like multicollinearity, feature redundancy, mediating behavior, and synergistic behavior \citep{covert2021explaining}.

Each individual observation in a dataset can be treated as a cooperative game, where the loss between the prediction and the ground truth is the reward function. The need for the Shapley properties is now clear: the overall dataset's Shapley value is the sum of the individual predictions, features that never change loss are given a zero value, identical players are handled correctly, and the Shapley values sum to the model's improvement in the loss function.

 Shapley values are highly conditional upon the choice of value function $\nu$. This choice of value function is critical to the analysis. For instance, it has been pointed out that the Shapley value function in SHAP can attribute positive value to covariates $X_i$ not used in the model due to the way it has been defined, whereas alternative choices of value function assign Shapley values of zero to covariates not used in the model \citep{chen2020true}.

Subset extensions and Shapley values form the basis of variable importance analysis. Most variable importance methods begin by defining some method for calculating  subset extensions $f_E$, and evaluate the effect of each feature's removal by calculating Shapley values of the change in the performance of $f_E$ under different feature subsets $s \in \mathbb{P}(p)$. The choice of how to generate the subset extension is also very important since there are many possible choices. Input-based methods that modify the data matrix $X$ have been known to generate impossible points that are not possible or vanishingly unlikely under the data-generating process $y \sim F(X) + \epsilon(X)$, which confuses the analysis by evaluating the model's performance on these impossible observations. Distribution and model-based methods need to be properly calibrated such that the distribution or model being used is accurate. And refit-based methods tend to be extremely computationally expensive \citep{covert2021explaining}.

\section{Selective Importance for Universal Approximators}

\textbf{Definition} Universal Approximator: a universal approximator is a class of functions $\mathbb{F}$ such that for any compact data $X$ and any arbitrary function $g(X)$, there exists some $f \in \mathbb{F}$ such that $sup_{x \in X}||f(x) - g(x)|| < \epsilon$ for any arbitrarily small $\epsilon$.

If the model class $\mathbb{F}$ is a universal approximator, and we have a data-generating process $y \sim F(X) + \epsilon(x)$, it is always possible for some $f \in \mathbb{F}$ to achieve the minimum possible loss such that $L(f(X), F(X)) = \mathbb{L}_0 = E(L(\epsilon(X)))$.

\textbf{Theorem} If a feature importance method produces a vector of weights $w$ that is selective with respect to the true data-generating process with respect to a class of functions $\mathbb{F}$ that is a universal approximator, the weights will be selective even if the model class is misspecified with respect to the true functional form.

For instance, if $\mathbb{F}$ is a class of tree models, which are universal approximators, but the true data-generating process can be expressed as a polynomial, any selective variable importance function on the class $\mathbb{F}$ will still identify the true set of covariates $X_T$ such that $F(X_T) = F(X) \forall X$.

We can prove this by contradiction. Assume that there is some $w_i \in w$ such that $w_i > 0$, yet $X_i \notin X_T$. This would then imply that $L(f(X), y) < L(f(X / X_i), y)$, assuming that $f$ is an optimal model that achieves the minimal possible loss. However, since $f \in \mathbb{F}$ is a model in a class of universal approximators, this cannot be true because then that would imply that $sup_{x \in X}||f(x) - g(x)|| > \epsilon$ for some cutoff $\epsilon$ value, increasing the loss.

\section{Refit-Based Importance for Selective Feature Importance}

Due to the aforementioned computational burden, most variable importance methods have focused on function-based replacements, rather than repeatedly refitting the entire model $f$. Refit-based importance methods tend to be especially costly when $f$ is a complex model.

However, we contend that a full refit is the only approach that is guaranteed to lead to a selective feature importance function. To see why, assume we have a subset extension function $f_E$ such that when $s_i = 0$, the values of $X_i$ are replaced or imputed in some way and passed back through the trained model $f$, whether by replacing values in $X_i$ with a constant value, distribution, or model output. We further know that by construction, $\exists X_i \ s.t. \ h_i(X, s) \neq X_i$ for any given $f_E$. In fact, some feature importance methods enforce "replaced" covariate columns to not be equal to the original column for all covariates. Then the change in the output between $f$ and $f_E$ can be represented as:

\[f(X) - f_E(X, s) = f(X) - f(X|\{X_i: s_i = 0 \to h_i(X, s)\})\]

Recall that we are concerned with the true DGP, $F(X) \sim y + \epsilon(X)$. Here, $\epsilon(X)$ constrains the minimal achievable loss on the given dataset. Denote this minimal loss as $E(L(F(X), y)) = E(L(\epsilon(X))) = \mathbb{L}_0$. 

Assume that the functional form of our trained model $f(X) \neq F(X)$, but for our desired loss function $L$, $L(f(X), y) = L(F(X), y) = \mathbb{L}_0$. This implies that the functional form of $f$ must use all the covariates in $X_T$, but may also use some unnecessary covariates not in $X_T$. Then the increased loss associated with replacing any given $X_i$ with $h_i(X, s)$ in our subset extension $f_E(X)$ is proportional to $L(f(X), y) - L(f(X|X_i = h_i(X, s)), y)$, assuming that $X_i$ is part of the functional form of $f$. 

We can prove that this quantity will be positive for all $X_i$ used in the functional form of $f$ under light assumptions. Assume that no column in $X$ is a perfect function of the other columns. This then implies that we cannot have $h_i(X, s) = X_i$ when $s_i = 0$, since no method (constant value, function based replacement, or distribution based replacement) can perfectly impute the values of $X_i$. We also know $L(f(X), y) = \mathbb{L}_0$, and by definition $f(X) \neq f_E(X, s) = f(X|X_i = h_i(X, s)))$. Therefore, predictions on the replaced values of $X_i$ cannot give the optimal loss $\mathbb{L}_0$. Therefore, we will always see increased loss for any $X_i$ used in the functional form of $f$ when using feature importance methods that adjust the inputs to a model $f$, but do not refit the model entirely.

If any $X_i$ is not part of the true covariate set $X_T$, but is part of the model $f$, then we have introduced a positive increase in the associated loss between $f_E(X, s)$ and $y$, without any assumptions as to the functional form of $h_i$. Therefore, unless $f$ happens to only use covariates in the true feature set $X_T$, we will see a decrease in model performance associated with the importance metric and therefore positive importance to the associated feature. Since these sorts of methods are not intended to be inferential about the true DGP, this is intended behavior, but it makes them less useful as inferential tools. Therefore, this class of feature importance methods cannot be guaranteed to give a selective feature importance method.

On the other hand, if $f_E(X, s)$ is the function $f$ refit without covariate $X_i$, then it is possible for $f(X) = f_E(X, s)$ if the only covariates removed are not in the true set $X_T$. Because functional replacement methods require the values of $X_i$ to be different from the true values post-removal while keeping the prediction model the same, the loss of the function post-replacement is necessarily higher. Refit based methods do not have this constraint if $X_i$ is not an element of $X_T$, because the minimal loss is still achievable even if $X_i$ is excluded by definition.

Note also that it is possible in an asymptotically large dataset to have $X_i \in X_f, X_i \notin X_T$, even when $L(f(X), y) = \mathbb{L}_0$. To see why, assume that $f$ is a gradient boosted decision tree model. Assume that $X = [X_1, X_2, X_3]$, and $y = X_1 + X_2 + \epsilon$. Let $X_3 = X_1 + X_2 + \epsilon + \gamma$, where $\gamma$ is an small added mean-zero noise term. This means $X_3$ is a noisy collinear combination of $X_1, X_2$.

The correlation $\rho(y, X_3) > max(\rho(y, X_2), \rho(y, X_1))$. This means that the gradient boosting algorithm will select the first split to be on $X_3$, as $X_3$ maximally separates different values of $y$ despite being a collinear combination and containing no marginal utility. Since the set of true variables is $X_T = [X_1, X_2]$, we have that $X_3 \in X_f, X_3 \notin X_T$ as desired. However, $X_3$ is not needed to achieve the optimal loss. A model fit only to $X_1, X_2$ will achieve $\mathbb{L}_0$. This implies a refit-based method could be guaranteed to correctly identify that $X_3$ is not an element of $X_T$.

\section{Conformal Inference and LOCO}

We now consider a specific approach to variable importance, where we assume that the model $f$ may be significantly mis-specified relative to the true model $F$. Since we make no assumptions as to the true functional form of $F$, we cannot resort to the common tools used for linear models. Instead, we turn to the tools of conformal inference.

\textbf{Definition} Leave-one-covariate-out (LOCO) Variable Importance: LOCO optimizes a full model $f$ and then estimates $p$ sub-models that are the model $f$ refit without each individual covariate $X_1...X_p$. This process can be repeated many times on a dataset using cross validation or bootstrap resamplings of $X$ to get a more consistent estimation of the marginal utility of each $X_i$ \citep{lei2018distribution}.

LOCO is a flexible approach. As defined above it makes no assumptions about the class of model $\mathbb{F}$ being used, or the distribution of the data $X$. Assuming that LOCO is run for several resampled iterations, conformal prediction bands for the change in the quality metric for each covariate $\nu(f(X)) - \nu(f(X / X_i))$ can be calculated. LOCO is also a selective variable importance method, as it fully refits the function $f$ at each step rather than using a single trained $f$. In an asymptotically large sample, this should be guaranteed to tell whether any given $X_i$ has a positive marginal importance to the overall feature set.

We focus on the global importance measures of LOCO. Assume that $\theta_j$ is the true value of $\nu(f(X)) - \nu(f(X / X_j))$, and $\hat{\theta_j}$ is the estimated value of $\theta_j$ from running $k$ iterations of LOCO. Then an asymptotic confidence interval for $\hat{\theta_j}$ is:

\[\hat{\theta_j} - \frac{\mathbb{z}_\alpha / s_j^2}{\sqrt{k/2}}, \hat{\theta_j} + \frac{\mathbb{z}_\alpha / s_j^2}{\sqrt{k/2}}\]

Where $s_j^2$ is the sample standard deviation. Then the following hypothesis test can be performed, using any one of a wide variety of nonparametric tests:

\[H_0: \theta_j > 0 \text{ versus } H_1: \theta_j \leq 0\]

The LOCO paper also suggests several alternative testing methodologies, such as using the median of the trials to estimate $\theta_j$ as a way to smooth out the distribution and allow for more standard tests, such as a Wilcoxon signed-rank test \citep{lei2018distribution}.

However, the LOCO methodology is missing one crucial element common to most variable importance functions. LOCO does not examine all possible sets of $s \in \mathbb{P}(p)$. Instead, it only examines the sets $[0, 1, ...1]$, $[1, 0, ... 1]$, ... $[1, 1, ... 0]$, which only constitute $p$ out of the $2^p$ possible sets. This means that LOCO is not estimating Shapley values of the variables with respect to any possible value function $\nu$. Instead, LOCO assumes that the set of $p$ feature subsets is a sufficient sample of the space of covariate subsets. Therefore, the Shapley value properties do not apply to LOCO importance values.

Specifically, LOCO values are not efficient when the variables $X_1...X_p$ are not fully independent. In this case, for any given variable $X_i$ that is correlated with other covariates, there is some marginal change in $\nu$ attributable to $X_i$ that is independent of all other $X_j, j \neq i$ which can be denoted $\xi_i$ and some dependent effect that is partially attributable to a set of covariates including $X_i$, denoted $\delta_i$. LOCO will only measure the independent $\xi_i$ effect associated with each variable, without measuring utility due to the $\delta_i$ between covariates. This can be a significant issue when a set of variables is highly collinear, and the values of $\delta_i$ are relatively large compared to the $\xi_i$. LOCO will always focus on each variable's marginal contribution, and miss the high utility of the overall group. 

LOCO values also violate symmetry when the variables are not independent. Consider a pair of variables, $X_i$ and $X_j$, such that $\xi_i = \xi_j$, but $\delta_i > \delta_i$. These variables will have identical LOCO scores, despite $X_i$ being more useful to smaller coalitions.

\section{Model Class Reliance}

\textbf{Definition} Model Class Reliance (MCR): MCR trains $k$ models $f_1...f_k$ and then evaluates the subset $f_{MCR} = [f_i: L(f_i(X), y) \leq \mathbb{L}_0 + \delta]$ for some positive constant $\delta$. MCR importance then calculates the exhaustive permutation importance within the set of $\delta$-optimal models $f_{MCR}$ to assess the feature importance. They refer to the set of $\delta$-optimal models as the Rashomon set, 

MCR exploits the varying functional forms of the models within this set of high-performing models, and assesses the importance of covariates to each model by testing the permutation importance with respect to every possible permutation of the columns. The advantage of this approach is that it does not rely on a single model like SHAP does, but the downside is that by not evaluating the validity of the diverse functional forms, only their overall loss, MCR is guaranteed to never capture the true DGP. There is only one truly correct DGP, therefore any average of multiple different explanations is necessarily wrong unless it could be guaranteed that the different explanations were all noisy yet unbiased estimators of the true DGP. To our knowledge, no such proof exists. 

More rigorously, we can show this using the properties of permutation importance and the construction of $f_{MCR}$. The permutation importance is defined as the decrease in a loss metric $L(f(X), y)$ when a feature $X_i$ is randomly permuted. Define this permutation to be $Perm(X, i)$ for the data matrix $X$ with $X_i$ permuted. Since MCR uses exhaustive permutation (all possible permutations of $X_i$, denoted $\bar{Perm(X, i)}$), the only way for a permutation to result in a decrease of $L(f(Perm(X, i)), y)$ is if the model is poorly fit, and would benefit from dropping $X_i$. By definition of the Rashomon Set, such a model could not be included unless the chosen $\delta$ was extremely large. Therefore, the only way for MCR to output a selective variable importance function is if every model in $f_{MCR}$ is using the correct feature set. Since MCR specifically seeks diverse functional forms, this seems incredibly unlikely in any real-world scenario.

MCR is also obviously not an attributive method, since it does not deal with any particular realized model. It is in a unique space, where it attributes the importance of different covariates to different possible models in the model class $\mathbb{F}$.

\section{Shapley Marginal Surplus for Strong Models}

We now introduce our method, Shapley Surplus for Strong Models (SMSSM). Fix a model class $\mathbb{F}$ and a maximal number of iterations $k$. We then sample $k$ elements of the power set of possible feature sets, $\mathbb{P}(p)$, evenly distributed across the possible number of nonzero covariates with at least two elements, $2...p$. Label these $k$ feature vector indicators as $S_{*, 1}...S_{*, k}$. For each $S_{*, i}$, we evaluate the cross-validation performance of models fit to that subset, and the marginal surplus of each feature within the subset as estimated by the cross-validation performance of the model without that feature. Finally, we restrict the top $b$ percent of the full models, and average the marginal contribution of each covariate to this top $b$ percent of models, where the cutoff loss can be denoted $L_b$. This is the Shapley value with respect to the following value function:

\[\nu(c) = L(f(c), y) I(L(f(c), y) \leq L_b)\]

This is similar to LOCO in approach. However, rather than only use a single model, we sample a space of many models to more precisely attribute marginal surplus to features using their Shapley values. The full algorithm can be written as follows:

\begin{algorithm}[H]
   \caption{Shapley Marginal Surplus for Strong Models}
   \label{alg:example}
\begin{algorithmic}
   \STATE {\bfseries Input:} covariates $X = [X_1...X_P]$, target $y$
   \STATE Initialize $k$ random subsets of the covariates, $S_{*, 1}...S_{*, k}$ of size $i\%p$ for $i = 1...k$
   \FOR{$i=1$ {\bfseries to} $k$}
   \STATE Cross-validate a model on $X \times S_{*, i}$ 
   \STATE Record the cross-validation performance $L_{S_{*, i}}$
   \FOR{$l=1$ {\bfseries to} $p$}
    \IF{$S_{l, i} = 1$:}
    \STATE Create $S_{i, X_l = 0}$ equal to $S_{*, i}$ except index $l = 0$
    \STATE Refit the model on $X \times s_{i, X_l = 0}$ 
    \STATE Record the cross-validation performance $L_{S_{*, i} / X_l}$
    \STATE Record the difference $\Delta_{i, l} = L_{S_{*, i} / X_l} - L_{S_{*, i}}$
    \ENDIF
   \ENDFOR
   \ENDFOR
   \STATE Calculate the top $b \%$ of the $L_{S_{*, i}}$
   \STATE Subset to the evaluations $L_{opt} = L_{S_{*, i}} \leq L_B$
   \STATE Calculate the Shapley values $\phi_1...\phi_p$ of the covariates $X_1...X_p$ using the $\Delta_{i, l}$ values corresponding to $L_{opt}$
   \STATE \textbf{return} $\phi_1...\phi_p$
\end{algorithmic}
\end{algorithm}

\begin{figure}[h]
    \centering
    \includegraphics[width=6cm]{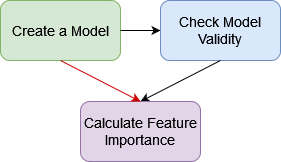}
    \caption{Most feature importance methods take the model as truth, going straight from creating a model to calculating importance (red arrow). Our procedure adds the intermediate step of calculating the validity of the class of models (black arrows), which dramatically improves feature importance allocation.}
\end{figure}

\textbf{Theorem} With an asymptotic sample size, SMSSM is a selective feature importance method with respect to the data-generating process when $L_B = \mathbb{L}_0$, assuming $k$ is also allowed to grow asymptotically.

This is straightforward from the definitions. First, observe that there is a fixed probability at each iteration that we select all of the elements of $X_T$. Assume that the number of members of the set is $T$ out of a total $p$ members, and at the current iteration we are going to have $sum(S_{*, i}) = j$. Then the probability of selecting the full set, plus some added covariates, is:

\[P_{T, j} = I(j \geq T) (\frac{j!}{T!})/(\frac{p!}{j!(p-j!)})\]

With an infinite sample size, any model that contains all members of $X_T$ will achieve $\mathbb{L}_0$. Similarly, in asymptotic sample the loss change associated with dropping a covariate $X_i \notin X_T$ will be zero. Therefore, only elements in $X_T$ will have nonzero $\Delta$ values, and therefore nonzero Shapley values, in these optimal loss samples.

\section{Methods and Data}

In this work, we primarily focused on simulated data, though we did use two real-world datasets. This is because only in simulated data can we have an explicit form for the data-generating process, and exactly evaluate to what extent the returned weights are selective with regard to this true data-generating process.

\subsection{Simulated Data}

We simulate five simple feature relationships where the ground truth is known. These are simple strong multicollinearity (DS 1), weak multivariate collinearity (DS 2), mediation (DS 3), a noisy polynomial (DS 4), multivariate strong collinearity (DS 5), and higher dimensional strong multicollinearity (DS 6). For further discussion of these datasets and the explicit data-generating processes used, please see the supplementary material.

\subsection{Real-World Data}

We cannot directly test the underlying accuracy with respect to a DGP of real data, since the real DGP is always unknown. However, similar to our argument for why MCR explanations are not selective, what we can assess is the consistency of calculated feature importances on a randomly split master dataset. If a method is not consistent on two evenly sized samples of a dataset, then one or both of the calculated importances is necessarily wrong. 

However, this is not an ideal test, because a high alignment between the two subsets is also not indicative of an accurate method; it is entirely possible for a method to be consistently wrong, meaning that the results can only be interpreted in the context of our results on simulated data with a known ground truth.

Datasets were pulled from the UCI machine learning repository. These include an Auto MPG dataset (Auto), a real estate valuation dataset (RE), a liver disease dataset (Liver), and a cancer prediction dataset (Cancer) \citep{misc_auto_mpg_9, misc_breast_cancer_14, misc_liver_disorders_60, misc_real_estate_valuation_477}. We evaluated our two selective methods, LOCO and SMSSM, on evenly split subsets of each dataset across five trials, and averaged performance.

\section{Results}

\subsection{Simulated Data Test}

We ran the full set of gradient boosted ensemble feature importance metrics for the  datasets described. The "XGB" metric is the gain score from the XGBoost model used for SHAP. For our method, we assessed the feature importance using the top 25\% of models.For simulated datasets 1, 5 and 6, we use the angle between the returned weights of the method and a known accurate ground truth weight. For simulated datasets 2, 3, and 4, we used the relative weight on the covariates not in $X_T$ as the score, where $X_T$ is known since the data was simulated according to a known DGP.  For the first metric, a low score is better; for the second, a high score is better. We denote these with the superscipt of 1 and 2 respectively on the table. These two metrics reference different important aspects of feature importance. In the first, we are trying to exactly match a ground truth. In the second, we are trying to see to what extent methods are selective, and can correctly allocate weight to true covariates.

\begin{table}[h]
\caption{Feature Importance Method Performance for Simulated Data}
\label{sample-table}
\vskip 0.15in
\begin{center}
\begin{small}
\begin{sc}
\begin{tabular}{lcccccr}
\toprule
- & DS$1^1$ & DS$2^2$ & DS$3^2$ & DS$4^2$ & DS$5^1$ & DS$6^1$\\
\midrule
XGB & 1.49 & 0.99 & 0.70 & 0.92 & 0.54 & 1.56\\
SHAP & 1.03 & 0.99 & 0.70 & 0.89 & 0.45 & 1.33\\
MCR & 0.97 & 0.99 & 0.74 & 0.92 & 0.46 & 1.28\\
LOCO & 0.12 & \textbf{1.00} & \textbf{1.00} & \textbf{1.00} & 0.62 & 0.64\\
SMSSM & \textbf{0.06} & 0.99 & \textbf{1.00} & 0.99 & \textbf{0.28} & \textbf{0.44}\\
\bottomrule
\end{tabular}
\end{sc}
\end{small}
\end{center}
\vskip -0.1in
\end{table}

For the simulated data, SMSSM emerges as highest average performer, and receives either the best or second best score on all six simulated datasets, though we must somewhat discount dataset 2 as all methods performed extremely well. Every other method dramatically underperforms on at least one dataset. The XGBoost feature importance and SHAP are consistent low performers, which is to be expected since these are attributive explanation tools, rather than selective. MCR is the worst performer of the methods that operate on multiple models, and does not stand out relative to LOCO and SMSSM. Despite excellent performance on datasets 3, and 4, LOCO is the worst performer among all methods on dataset 5, and also does not perform as well as SMSSM on datasets 1 and 6. This is because the variables in these datasets have an extremely high degree of collinearity, and LOCO does not handle interdependent importance correctly, instead only evaluating each feature's marginal utility. This leads to under-performance on highly collinear tasks, as evidenced by LOCO receiving the worst score on dataset 5.

\subsection{Real Data Results}

We also present the results of alignment between two randomly selected subsets of real datasets. As with datasets 1 and 5 in the simulated data, a lower score is better for this metric, as we also used the vector angle method to evaluate performance.

\begin{table}[H]
\caption{Average Score Over Five Trials for Real Data}
\label{sample-table}
\vskip 0.15in
\begin{center}
\begin{small}
\begin{sc}
\begin{tabular}{lccccr}
\toprule
Dataset & LOCO & SMSSM\\
\midrule
Auto & \textbf{0.249} & 0.264 \\
RE  & 0.721 & \textbf{0.504} \\
Liver & 0.786 & \textbf{0.705} \\
Cancer & 1.111 & \textbf{0.596} \\
\midrule
Mean & 0.717 & \textbf{0.517}\\
\bottomrule
\end{tabular}
\end{sc}
\end{small}
\end{center}
\vskip -0.1in
\end{table}

From the above, we can see that SMSSM is also the most consistent of our two selective feature importance methods, achieving a lower angle between trials on paired datasets in three out of four trials. This reinforces our hypothesis that SMSSM is likely to be giving more accurate feature importance scores, as lower consistency implies a less trustworthy feature importance function. We believe this is due to the added noisiness LOCO incurs by only focusing on each feature's marginal utility, at the cost of the added robustness of analyzing the mutual utility between covariates as well.

\section{Discussion}

\begin{table}[H]
\caption{Average Rankings of Feature Importance Methods}
\label{sample-table}
\vskip 0.15in
\begin{center}
\begin{small}
\begin{sc}
\begin{tabular}{lccccr}
\toprule
Method & Average Rank & Best & Worst\\
\midrule
XGB & 4.33 & 3 & 5 \\
SHAP & 3.67 & 2 & 5\\
MCR & 3.17 & 3 & 4\\
LOCO & 2 & \textbf{1} & 5\\
SMSSM & \textbf{1.17} & \textbf{1} & \textbf{2}\\
\bottomrule
\end{tabular}
\end{sc}
\end{small}
\end{center}
\vskip -0.1in
\end{table}

We calculate the average performance of each method on the simulated data. From the relative performances, a clear winner emerges. SMSSM has both the best overall performance as well as the most consistent performance, ranking as either the best or second best method on every dataset. 

Three methods - MCR, LOCO, and SMSSM have extremely high compute budgets required to make them work. If a faster method is required, the vanilla XGBoost importance metric is the best. SHAP does a good job of explaining models, but explaining nonlinear models is frequently harmful for actually explaining the true DGP as shown above.

The most interesting approach is LOCO, which did well on four simulated datasets but was the lowest performer on the fifth simulated dataset, and significantly underperformed SMSSM on dataset 1 despite being the second best method. The common factor is high multicollinearity between covariates. Since LOCO drops covariates one at a time, rather than in groups like SMSSM, it appears to struggle when multicollinearity is high. This provides a potentially interesting extension to LOCO where a grouped LOCO approach might work well in finite samples, where multicollinearity between predictors can throw off estimates.

Overall, we believe this is a valuable contribution to the model-based feature importance literature. We have demonstrated the limits of model-based explanations, and demonstrated that sampling the space of possible models is the more effective approach.

\section{Future Work}

The most obvious future direction for SMSSM is to incorporate a meta-model to estimate the optimal feature set. The performance of the overall model at each time step could be treated as a multi-armed bandit problem, with the inclusion or exclusion effect of each covariate as the target variable. 

This approach would provide both theoretical and practical benefits. In a theoretical sense, sampling more of the space of highly promising models and less of the space of unpromising models should decrease the noise of the procedure and improve the convergence rate. In a practical sense, this would also provide a large computational benefit, as more efficient sampling would require fewer overall samples while delivering the same or higher-quality results.

This is important because currently, the biggest bottleneck in using higher-quality importance methods like LOCO or SMSSM is the compute required, making these methods infeasible in large sample. 

Otherwise, we feel that this work is a promising first step towards inferential explanations for machine learning methods, that both respect the model used as well as the underlying DGP. If these methods could be made more efficient, it might someday be possible to use powerful and flexible models in the same way linear models are used today.

\bibliography{sample_paper}

\begin{thebibliography}{14}
\providecommand{\natexlab}[1]{#1}
\providecommand{\url}[1]{\texttt{#1}}
\expandafter\ifx\csname urlstyle\endcsname\relax
  \providecommand{\doi}[1]{doi: #1}\else
  \providecommand{\doi}{doi: \begingroup \urlstyle{rm}\Url}\fi

\bibitem[mis(1990)]{misc_liver_disorders_60}
{Liver Disorders}.
\newblock UCI Machine Learning Repository, 1990.
\newblock {DOI}: https://doi.org/10.24432/C54G67.

\bibitem[Branzei et~al.(2008)Branzei, Dimitrov, and Tijs]{branzei2008models}
Branzei, R., Dimitrov, D., and Tijs, S.
\newblock \emph{Models in cooperative game theory}, volume 556.
\newblock Springer Science \& Business Media, 2008.

\bibitem[Burkart \& Huber(2021)Burkart and Huber]{burkart2021survey}
Burkart, N. and Huber, M.~F.
\newblock A survey on the explainability of supervised machine learning.
\newblock \emph{Journal of Artificial Intelligence Research}, 70:\penalty0 245--317, 2021.

\bibitem[Chen et~al.(2020)Chen, Janizek, Lundberg, and Lee]{chen2020true}
Chen, H., Janizek, J.~D., Lundberg, S., and Lee, S.-I.
\newblock True to the model or true to the data?
\newblock \emph{arXiv preprint arXiv:2006.16234}, 2020.

\bibitem[Covert et~al.(2021)Covert, Lundberg, and Lee]{covert2021explaining}
Covert, I., Lundberg, S., and Lee, S.-I.
\newblock Explaining by removing: A unified framework for model explanation.
\newblock \emph{Journal of Machine Learning Research}, 22\penalty0 (209):\penalty0 1--90, 2021.

\bibitem[Fisher et~al.(2019)Fisher, Rudin, and Dominici]{fisher2019all}
Fisher, A., Rudin, C., and Dominici, F.
\newblock All models are wrong, but many are useful: Learning a variable's importance by studying an entire class of prediction models simultaneously.
\newblock \emph{J. Mach. Learn. Res.}, 20\penalty0 (177):\penalty0 1--81, 2019.

\bibitem[Kumar et~al.(2020)Kumar, Venkatasubramanian, Scheidegger, and Friedler]{kumar2020problems}
Kumar, I.~E., Venkatasubramanian, S., Scheidegger, C., and Friedler, S.
\newblock Problems with shapley-value-based explanations as feature importance measures.
\newblock In \emph{International Conference on Machine Learning}, pp.\  5491--5500. PMLR, 2020.

\bibitem[Lei et~al.(2018)Lei, G’Sell, Rinaldo, Tibshirani, and Wasserman]{lei2018distribution}
Lei, J., G’Sell, M., Rinaldo, A., Tibshirani, R.~J., and Wasserman, L.
\newblock Distribution-free predictive inference for regression.
\newblock \emph{Journal of the American Statistical Association}, 113\penalty0 (523):\penalty0 1094--1111, 2018.

\bibitem[Lundberg \& Lee(2017)Lundberg and Lee]{lundberg2017unified}
Lundberg, S.~M. and Lee, S.-I.
\newblock A unified approach to interpreting model predictions.
\newblock \emph{Advances in neural information processing systems}, 30, 2017.

\bibitem[Lundberg et~al.(2020)Lundberg, Erion, Chen, DeGrave, Prutkin, Nair, Katz, Himmelfarb, Bansal, and Lee]{lundberg2020local}
Lundberg, S.~M., Erion, G., Chen, H., DeGrave, A., Prutkin, J.~M., Nair, B., Katz, R., Himmelfarb, J., Bansal, N., and Lee, S.-I.
\newblock From local explanations to global understanding with explainable ai for trees.
\newblock \emph{Nature machine intelligence}, 2\penalty0 (1):\penalty0 56--67, 2020.

\bibitem[Quinlan(1993)]{misc_auto_mpg_9}
Quinlan, R.
\newblock {Auto MPG}.
\newblock UCI Machine Learning Repository, 1993.
\newblock {DOI}: https://doi.org/10.24432/C5859H.

\bibitem[Shapley(1953)]{shapley1953value}
Shapley, L.
\newblock A value for n-person games.
\newblock 1953.

\bibitem[Yeh(2018)]{misc_real_estate_valuation_477}
Yeh, I.-C.
\newblock {Real Estate Valuation}.
\newblock UCI Machine Learning Repository, 2018.
\newblock {DOI}: https://doi.org/10.24432/C5J30W.

\bibitem[Zwitter \& Soklic(1988)Zwitter and Soklic]{misc_breast_cancer_14}
Zwitter, M. and Soklic, M.
\newblock {Breast Cancer}.
\newblock UCI Machine Learning Repository, 1988.
\newblock {DOI}: https://doi.org/10.24432/C51P4M.

\end{thebibliography}
\bibliographystyle{icml2023}

\end{document}